\documentclass[sigconf]{acmart}

\usepackage{times}
\usepackage{soul}
\usepackage{url}
\usepackage{graphicx}
\usepackage{amsmath}
\usepackage{amsthm}
\usepackage{booktabs}
\usepackage{algorithm}
\usepackage{algorithmic}
\usepackage[switch]{lineno}
\usepackage{multirow}
\usepackage{float}
\usepackage{colortbl}
\usepackage{textcomp}     
\usepackage{pifont}
\usepackage[table]{xcolor}

\hypersetup{
  hidelinks,
  pagebackref=true,
  breaklinks=true,
  colorlinks=true,
  linkcolor=blue,
  citecolor=blue,
  urlcolor=blue
}

\begin{document}

\title{SpatialLock: Precise Spatial Control in Text-to-Image Synthesis}



\author{Biao Liu}
\affiliation{%
  \institution{The Sugon Group}
  \city{Xi’an}
  \country{China}}
\email{lbiao7428@gmail.com}

\author{Yuanzhi Liang}
\affiliation{%
  \institution{TeleAI, China Telecom}
  \city{Shanghai}
  \country{China}}
\email{liangyzh18@outlook.com}


\begin{abstract}
  Text-to-Image (T2I) synthesis has made significant advancements in recent years, driving applications such as generating datasets automatically. However, precise control over object localization in generated images remains a challenge. \textbf{\textit{Existing methods fail to fully utilize positional information, leading to an inadequate understanding of object spatial layouts.}} To address this issue, we propose \textit{SpatialLock}, a novel framework that \textit{\textbf{leverages perception signals and grounding information to jointly control the generation of spatial locations.}} \textit{SpatialLock} incorporates two components: \textbf{Po}sition-Engaged \textbf{I}njection (\textit{PoI}) and \textbf{Po}sition-\textbf{G}uided Learning (\textit{PoG}). \textit{PoI} directly integrates spatial information through an attention layer, encouraging the model to learn the grounding information effectively. \textit{PoG} employs perception-based supervision to further refine object localization. Together, these components enable the model to generate objects with precise spatial arrangements and improve the visual quality of the generated images. Experiments show that \textit{SpatialLock} sets a new state-of-the-art for precise object positioning, achieving IOU scores above $0.9$ across multiple datasets, as shown in \hyperref[fig:fig1]{Figure 1}.
\end{abstract}



\begin{teaserfigure}
    \centering
  \includegraphics[width=1\textwidth]{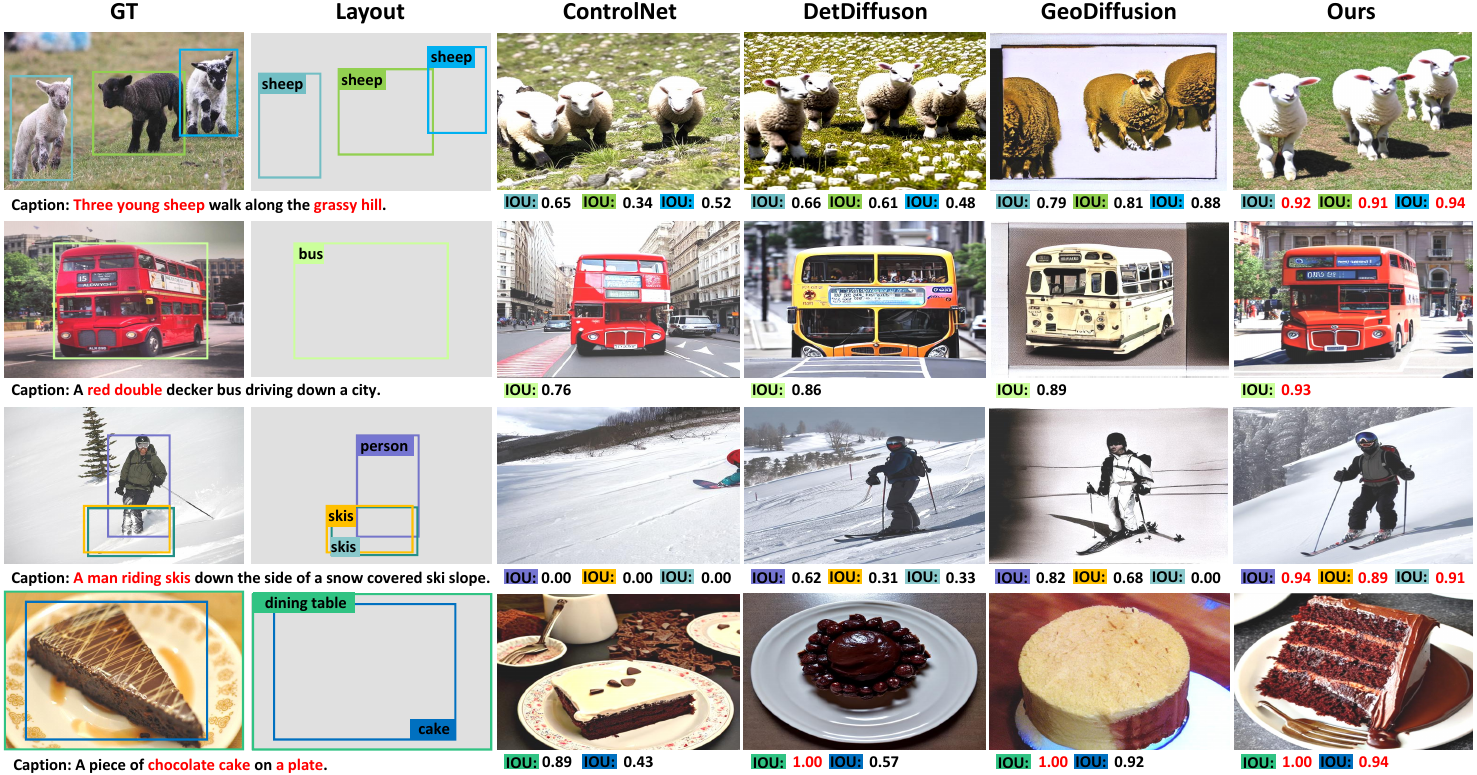}
  \caption{In various scenarios, \textit{SpatialLock} generates vivid images with more accurate position object entities compared to other methods, demonstrating the efficacy of our approach (we assess the accuracy of the generated object entities using the IOU, and the best performance is highlighted in \color{red} red\color{black}).}
  \label{fig:fig1}
\end{teaserfigure}


\maketitle

\vspace{-0.5em}
\section{Introduction}
Generative models have advanced significantly in recent years, particularly Diffusion Models (DMs), which exhibit exceptional generation quality due to their unique mechanisms and robust generation processes \cite{ramesh2022hierarchical,saharia2022photorealistic,rombach2022high}. These capabilities have unlocked numerous new possibilities, such as generating high-quality datasets \cite{wang2017effectiveness,bowles2018gan}, addressing data limitations in machine learning \cite{wang2016machine}, and enabling large-scale model training \cite{narayanan2021efficient}.

\begin{figure}[!h]
    \centering
    \includegraphics[width=1\linewidth]{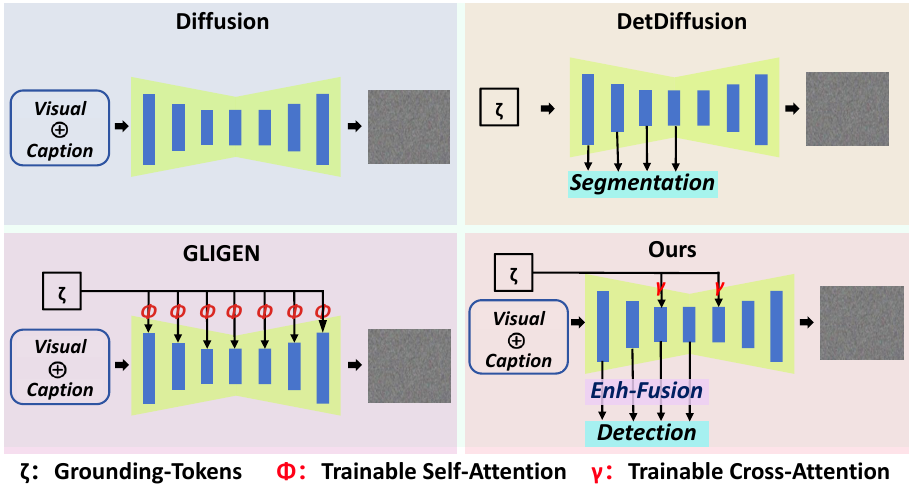}
    \caption{Compared to prior methods, our method uses cross-attention and a perception module to ensure the sufficient encoding of location information, additionally, we incorporate \textit{Enh-Fusion} into the perception module to further integrate global information.}
    \label{fig:fig2}
\end{figure}

However, despite their high fidelity and resolution, DMs often struggle with precise control over object localization \cite{jiang2024pixelman}. The inherent lack of coordinate-level positional input in original DMs leads to generated instances frequently deviating from intended locations. Although recent efforts have integrated position signals as input to improve DMs’ positional control \cite{li2023GLIGEN,wang2024DetDiffusion,chen2023geodiffusion,cheng2023layoutdiffuse}, \textbf{\textit{existing methods still underutilize the position information.}} Specifically, as shown in \hyperref[fig:fig2]{Figure 2}, some approaches lack explicit guidance for location information \cite{li2023GLIGEN}, while others fail to empower specific understanding ability for the position \cite{wang2024DetDiffusion}. This limits their ability to fully encode spatial features, hindering the generation of images with accurate object layouts.

To address the challenges of precise object localization, we propose \textit{SpatialLock}, a novel T2I framework that \textbf{\textit{achieves accurate spatial arrangements}} \textit{\textbf{through the integration of grounding and perception information.}} \textit{SpatialLock} extends a base T2I model \cite{rombach2022high} by introducing two components: Position-Engaged Injection (\textit{PoI}) and Position-Guided Learning (\textit{PoG}). These components enhance the model's ability to learn spatial information and generate objects at specified locations. Specifically, \textit{PoI} provides a mechanism to incorporate grounding information directly into the network during training, while \textit{PoG} introduces perception supervision to further refine spatial accuracy. Together, these components improve localization precision and preserve the high fidelity of generated images, allowing the model to deliver consistent and visually accurate results.

Next, we describe the proposed methods in detail. \textit{PoI} introduces an attention module, Grounding-Attention  \cite{vaswani2017attention}, to process grounding information via Grounding-Tokens \cite{li2023GLIGEN}, enabling the model to effectively integrate spatial information with visual and textual features. This module enhances the network's capacity to learn positional relationships and ensures more accurate placement of object entities. On the other hand, \textit{PoG} provides additional supervision by utilizing perception information, which guides the network to adhere strictly to the input spatial layout, and \textbf{\textit{to enhance contextual understanding and mitigate the interference of perception information on the diffusion process}}, we integrate \textit{Enh-Fusion}, a Transformer-based fusion module with stronger representation capabilities, into the \textit{PoG}. Leveraging these constraints on spatial information, \textbf{\textit{SpatialLock achieves a deeper understanding of the semantic information in the input data, thereby improving the positional accuracy of generated objects.}}

Experiments demonstrate that \textit{SpatialLock} achieves state-of-the-art performance in generating object entities with precision location. For instance, on the MSCOCO dataset \cite{lin2014microsoft}, \textit{SpatialLock} achieved a $32.1$ mAP, outperforming previous methods by $0.9$, and demonstrating a $1.34$ improvement in FID. These results highlight the effectiveness of integrating grounding and perception signals into T2I models, enabling precise object localization with fewer trainable parameters.

The primary contributions of our work are as follows:

1. We propose\textit{ SpatialLock}, a novel T2I framework that integrates grounding and perception information to achieve precise object localization in generated images.

2. We design and introduce two components, \textit{PoI} and \textit{PoG}, which enhance the ability of the model to learn positional relationships and refine spatial layouts through grounding and perception supervision, respectively.

3. \textit{SpatialLock} establishes new state-of-the-art performance in image generation tasks, achieving superior results in localization precision and image fidelity.

\begin{figure*}[!ht]
    \centering
    \includegraphics[width=0.8\linewidth]{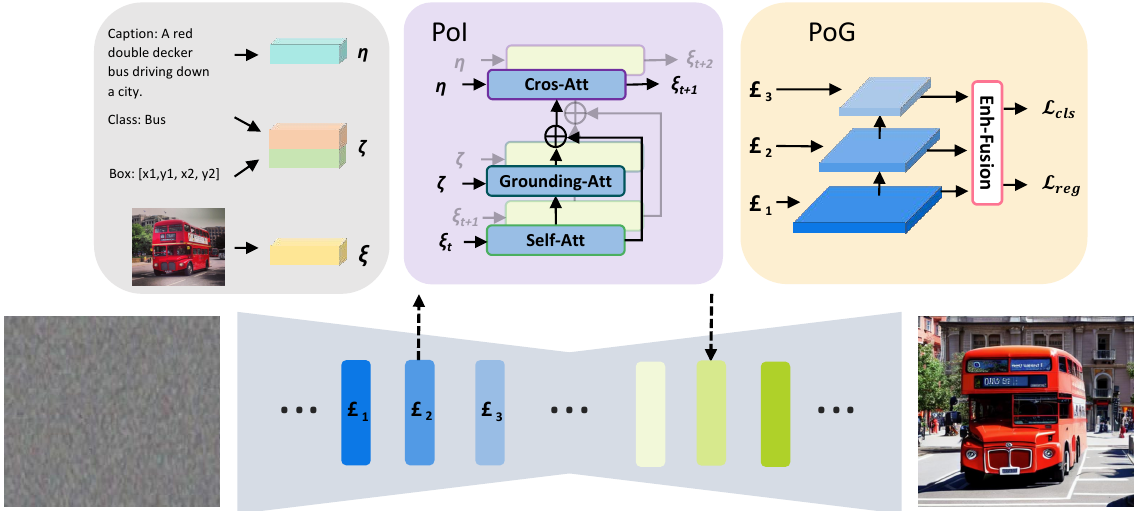}
    \caption{\textit{SpatialLock} incorporates \textit{PoI} and \textit{PoG} into the original Stable Diffusion framework. \textit{PoI} learns grounding information directly through Grounding-Attention to generate the precise positional layout, and \textit{PoG} provides an extra supervisor by perception information to refine the object location.}
    \label{fig:fig3}
\end{figure*}

\section{Related Work}
\paragraph{\textbf{Diffusion Models.}} Diffusion models (DMs) achieve image generation by learning the bidirectional transition between images and Gaussian noise \cite{ramesh2022hierarchical,saharia2022photorealistic,rombach2022high}. This unique learning mechanism results in superior fidelity and image quality compared to traditional GANs \cite{goodfellow2014generative,karras2019style,karras2020analyzing}. Due to their outstanding performance, DMs have garnered significant research interest, with various domains exploring their application to alleviate repetitive tasks \cite{stockl2023evaluating,hong2024massive}. Some researchers have integrated DMs with latent spaces, creating a transformative architecture known as Latent Diffusion Models (LDMs) \cite{kwon2022diffusion}, which substantially improves training speed and generation efficiency. However, the original LDMs primarily control generation results through text-captions, limiting their generative capabilities \cite{dhariwal2021diffusion}. To address this issue, we inject position information into the LDM framework, enabling the network to control the position of generated objects and thus address a broader spectrum of generation requirements.
\paragraph{\textbf{Generated Image with Layout.}} Generating images with specified layouts has always been a focal point of research, involving the synthesis of corresponding images given object bounding boxes and categories \cite{cheng2023layoutdiffuse,zhao2019image}. Over the past years, notable advancements have been achieved \cite{gupta2021layouttransformer,kong2022blt}. Earlier approaches, based on convolutional neural networks (CNNs), made significant strides \cite{zhao2019image}, with those methods primarily relying on auto-encoders (VAEs) \cite{masci2011stacked}, which require fewer training resources. Subsequently, the exploration of Transformer architectures \cite{vaswani2017attention} has led to an increasing number of methods leveraging these mechanisms. GLIGEN \cite{li2023GLIGEN} integrates self-attention layers into standard LDMs to extract grounding information, while LayoutDiffuse \cite{cheng2023layoutdiffuse} adopts a customized Transformer module to encode bounding box information. GeoDiffusion \cite{chen2023integrating} uses grounding, rather than text-captions, to control object positioning. Despite these advancements, most of these approaches focus on integrating position information through the Transformer itself, which leads to insufficient supervision of the network and results in an imprecise layout. DetDiffusion explores the impact of perception information on layouts \cite{wang2024DetDiffusion}. However, it lacks a dedicated module for explicit layout learning. These limitations hinder the effective encoding of position information from input data.

\section{Method}

To enhance the ability of Text-to-Image (T2I) models to learn positional features \cite{wang2024DetDiffusion}, we propose two components: Position-Engaged Injection (\textit{PoI}) and Position-Guided Learning (\textit{PoG}). \textit{\textbf{PoI introduces grounding information via  Ground-Attention, enabling the effective fusion of spatial data with other features.}} \textbf{\textit{PoG provides additional perception supervision during training, strengthening the network’s ability to refine locational features.}} Together, these components improve the interaction between spatial and generative information, resulting in precise object layouts. The overall architecture is depicted in \hyperref[fig:fig3]{Figure 3}, and the details of \textit{PoI} and \textit{PoG} are described in Sections 3.1 and 3.2, respectively.

\subsection{Position-Engaged Injection}
Original Text-to-Image (T2I) models lack a dedicated module for explicitly learning position features, which prevents the network from directly receiving coordinate-level location information, resulting in inaccurate spatial placements of generated objects \cite{rombach2022high}. To address this limitation, \textbf{\textit{we propose the \textit{PoI} to integrate visual elements, captions, and position information into the T2I framework, thereby enabling precise control over object locations}}.

\paragraph{\textbf{Input-Tokens.}} For the precision of object localization, Grounding-Tokens \cite{li2023GLIGEN}, incorporating position information via bounding boxes $\boldsymbol{b}$ and category labels $\boldsymbol{c}$, are introduced to improve the learning of location. The bounding box $\boldsymbol{b}$, defined by the top-left and bottom-right coordinates, is encoded using Fourier embeddings \cite{li2023embedding}, while category information $\boldsymbol{c}$ is encoded similarly to text-caption tokens \cite{radford2021learning}. These features are then fused using a Multi-Layer Perceptron (MLP) \cite{riedmiller2014multi}, concatenating the two inputs across the feature dimension, to produce a grounding feature, as $\boldsymbol{\zeta}$ shown in \hyperref[fig:fig3]{Figure 3}. Finally, the grounding tokens, along with visual features and caption tokens, are concatenated into a unified input $\boldsymbol{I}$:
\begin{equation}
    \boldsymbol{I} = (\boldsymbol{V}, \boldsymbol{C}, \boldsymbol{G}),
\end{equation}
\begin{equation}
    \boldsymbol{G} = \text{MLP}(f_{text}(\boldsymbol{c}), \text{Fourier}(\boldsymbol{b})),
\end{equation}

where $\boldsymbol{G}$ is grounding information denoted as [$G_1$, $G_2$, $\dots$ , $G_N$] and the $N$ is the number of objects, $f_{text}$ represents text encoding. $\boldsymbol{V}$ is a visual feature obtained by a pre-trained Vector Quantized Variational AutoEncoder \cite{wang2016auto}, and $\boldsymbol{C}$ is a caption feature acquired via a pre-trained CLIP text encoder \cite{radford2021learning}.

Unlike prior works, such as DetDiffusion \cite{wang2024DetDiffusion}, \textbf{\textit{we retain the text-captions as input, allowing it to influence the overall layout while blending it with grounding information through cross-attention}}, which enhances the network’s ability to generate objects with accurate location while improving the fidelity of generated images.

\paragraph{\textbf{Grounding-Attention.}} To incorporate grounding information, we design a trainable Transformer module \cite{li2023GLIGEN,vaswani2017attention}, which is inserted between the original self-attention and cross-attention layers of the T2I model \cite{rombach2022high}. This module utilizes $\boldsymbol{G}$ and $\boldsymbol{V}$ as input to facilitate the fusion of visual and grounding features. To preserve the ability to generate high-quality images guided by text-captions, we retain the original structure of the T2I network and freeze the parameters during training. The Grounding-Attention is expressed as:
\begin{equation}
    \boldsymbol{v} = \boldsymbol{v} + \tanh(\gamma) \cdot \text{CrossAttn}(\boldsymbol{V}, \boldsymbol{G}),
\end{equation}
where $\gamma$ is a learnable parameter initialized to 0, dynamically regulating the influence of grounding signals to preserve image fidelity.

To optimize efficiency, Grounding-Attention is applied selectively in the last Down-sampling and first Up-sampling layers of the U-Net \cite{wang2022unetformer}. \textbf{\textit{These layers contain rich semantic features, making them effective for capturing spatial layouts and exploring the relationship between the position and visual elements.}} This selective insertion reduces the number of trainable parameters and significantly conserves training resources. 

Prior approaches, such as GLIGEN \cite{li2023GLIGEN}, treated captions and grounding as a unified entity, employing a specialized self-attention module for information capture. In contrast, \textbf{\textit{we treat them as separate components, leveraging cross-attention to blend their features, which mitigates the potential for rich semantic details within visual information to obscure relatively sparse positional features}} \cite{zhao2022word2pix}. This strategy minimizes the influence of visual information on location features, which more effectively generates objects with precise locations.

\subsection{Position-Guided Learning}

While \textit{PoI} integrates grounding information, \textbf{\textit{PoG}} \textit{\textbf{focuses on perception supervision to refine the spatial location}}. This component introduces a perception module consisting of classification and regression branches to guide the T2I network in learning spatial features during training \cite{zou2023object}. \textbf{\textit{Classification features facilitate the generation of accurate objects, while regression information aids in determining a better spatial layout.}} Furthermore, studies show that Transformer modules are prone to catastrophic forgetting, where they blur previously learned knowledge as new information is added \cite{kotha2023understanding}. \textit{PoI} is designed to refine position features, thereby optimizing this problem.

\paragraph{\textbf{Feature from U-Net.}} Building upon studies that have validated the exceptional performance of U-Net architectures in perception tasks \cite{hertz2022prompt,wu2023datasetdm}, \textit{PoG} extracts multi-scale features from U-Net’s Down-sampling layers for classification and regression. In particular, these layers encode rich semantic information, which is highly beneficial for perception networks, and features from different scales improve the receptive field and provide diverse spatial information. These features are then used as the input of the Feature Pyramid Network (FPN) to further fusion \cite{lin2017feature}.

\paragraph{\textbf{Enhance Fusion.}} Owing to the constrained receptive field and the lack of global context representations in CNN-based perception information, inserting the perception module into diffusion models can adversely affect generative performance \cite{tian2024dic, shang2024resdiff}. To mitigate this issue, we incorporate a multi-head Transformer module into the \textit{PoG} component, which takes features from the FPN \cite{lin2017feature} as input and leverages a self-learned positional embedding \cite{zhu2020deformable} to facilitate effective spatial integration. This design not only addresses the contextual deficiencies of CNN-based architectures but enables a deeper fusion of position and generation information, resulting in more coherent feature representations that better guide the diffusion process. The above information fusion process can be formally expressed as:

\begin{equation}
\boldsymbol{X} = \text{Concat}(\boldsymbol{F_{\text{fpn}}}) + PE, \quad \boldsymbol{X} \in \mathbb{R}^{(H \cdot W) \times D},
\end{equation}

\begin{equation}
\boldsymbol{X'} = \text{MSA}(\boldsymbol{X}) = \text{Softmax}\left( \frac{QK^\top}{\sqrt{d_k}} \right)V,
\end{equation}

\begin{equation}
\boldsymbol{F_{\text{Enh}}} = \text{Reshape}(\boldsymbol{X'}) \in \mathbb{R}^{C' \times H \times W},
\end{equation}
where $PE$ is a learnable positional embedding. $\boldsymbol{F_{Enh}}$ denotes the contextually enhanced feature after fusion.

\paragraph{\textbf{Perceptual Network.}} The perception module comprises two independent branches: 1. Classification Branch: Processes U-Net features through four convolutional layers to predict object categories. 2. Regression Branch: Uses a similar structure to predict bounding box coordinates. In \textit{PoG}, a decoupled architecture, separating the classification and regression branches, is adopted to mitigate the mismatch between these branches, and an anchor point strategy is employed to effectively handle positive and negative samples. Finally, the perception features will be integrated with the generation information to enhance the network’s learning for spatial layouts.

The perception module introduces a loss combined classification loss $L_{cla}$ derived from Cross-Entropy and regression loss $L_{reg}$ derived from Focal Loss \cite{ross2017focal}, which can be expressed as:
\begin{equation}
    \boldsymbol{L_P} = \boldsymbol{L_{cla}} + \boldsymbol{L_{reg}},
\end{equation}

The final loss function combines perception loss $L_P$ and generation loss $L_{SD}$: 
\begin{equation}
    \boldsymbol{L} = \alpha \boldsymbol{L_P} + \beta \boldsymbol{L_{SD}},
\end{equation}
where $\alpha$ and $\beta$ are learnable parameters that balance the contributions of perception and generation information.

\begin{figure*}[!ht]
    \centering
    \includegraphics[width=1\linewidth]{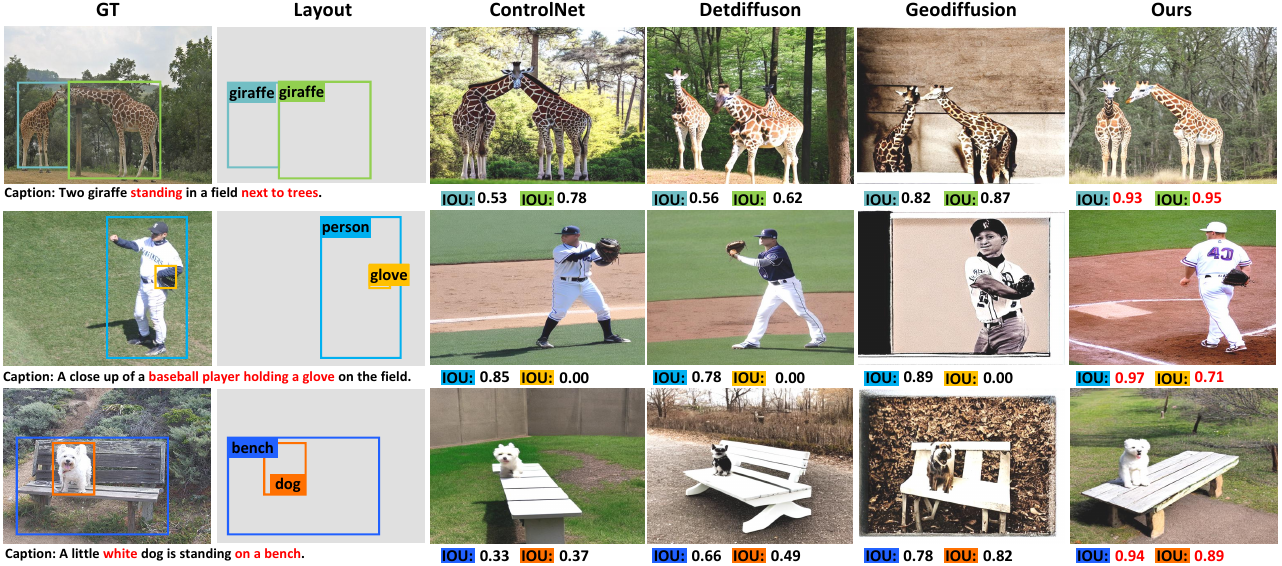}
    \caption{Compared to other advanced methods on MSCOCO, our model produces more realistic images with superior spatial layout, especially for small objects (e.g., the correctly placed \textbf{baseball glove}). Note that the performance of DetDiffusion is the result of our reproduction and may differ from the original.}
    \label{fig:fig4}
\end{figure*}

\section{Experiments}

Due to space limitations, additional results and the ablation study of $\alpha$ and $\beta$ are provided in the Appendix.
\subsection{Evaluation Datasets and Metrics}
\paragraph{\textbf{Datasets:}}
We train and evaluate our model using two datasets: Flickr \cite{hopken2020flickr} and MSCOCO \cite{lin2014microsoft}, and with all samples annotated in English. The Flickr dataset comprises $30,000$ images, each with annotations including bounding boxes, key points, and text captions. Specifically, each sample includes five captions, with each caption consisting of tokens representing individual words. The MSCOCO dataset, a widely utilized benchmark, consists of $118,287$ training images and $5,000$ validation images. Each image in MSCOCO is annotated with bounding boxes, object categories, and captions.

\paragraph{\textbf{Metrics:}}
We utilize mean Average Precision (mAP) \cite{yue2007support} to evaluate the accuracy of generated object positions. To ensure the accuracy of the assessment, we retrained a detector based on the original YOLOv4 \cite{jocher2022ultralytics}, aligning its input with the output of our generation model, which mitigates accuracy discrepancies arising from variations in feature dimensions. Additionally, the Fréchet Inception Distance (FID) \cite{yu2021frechet} is employed to quantify the fidelity of the generated images. FID measures similarity by calculating the Fréchet distance between generated and real image distributions, a method widely recognized for assessing image fidelity.

\subsection{Implementation Details}
The subsequent section details our experiment specifics. To maintain the capability for generation, we fine-tune our model based on the Stable Diffusion v1.4 checkpoint \cite{ramesh2022hierarchical,saharia2022photorealistic,rombach2022high}, and we trained our model using 8 NVIDIA V100 (32GB) GPUs for a total of 28 hours.

\paragraph{\textbf{Extract Feature:} }
Our model, unlike the original Stable Diffusion, incorporates Grounding-Tokens to expand the input features specifically for learning object location. We will now detail the method for extracting these features.
For text-captions encoding, we employ CLIP \cite{radford2021learning}, a robust representation method introduced by OpenAI in 2021 that has found widespread adoption in visual captioning tasks. CLIP provides a $77\times768$ dimensional tensor representation of the caption feature. In our approach, we utilize CLIP solely for caption feature extraction without parameter fine-tuning.

We encode the input image into a latent space using a Variational Autoencoder (VAE) \cite{wang2016auto}. The resulting latent feature has dimensions of $64\times64$, a significant reduction compared to the original image size. Consistent with our approach to the CLIP embedding, the VAE parameters remain fixed during training.

Grounding-Tokens are composed of categorical text tokens [$cat$, $person$, $dog$, $car...$] and bounding box coordinates [$\alpha\_{min}$, $\beta\_{min}$, $\alpha\_{max}$, $\beta\_{max}$]. We employ a method analogous to text-captions processing to extract category features, yielding a $30\times768$ dimensional vector. For the bounding box, we apply Fourier Embedding \cite{li2023embedding} to encode these coordinates, resulting in a $30\times768$ feature map. A multi-layer perceptron (MLP) \cite{riedmiller2014multi} is then used to concatenate the categories and bounding box tensors. The output of MLP is a $60 \times 768$ tensor, which effectively integrates the information from both feature sets.

\paragraph{\textbf{Fuse Feature:}}
Our network accepts visual data, text-captions, and Grounding-Tokens as input. These features are progressively integrated into the model via a carefully designed strategy.

Within U-Net’s Down-sampling and Up-sampling layers \cite{sha2021transformer,wang2022unetformer}, self-attention modules capture intra-visual correlations. Cross-attention then integrates visual and textual features to assess their impact on image quality. To further refine feature fusion, Grounding-Attention, operating on both visual features and Grounding-Tokens, is strategically incorporated between self-attention and cross-attention layers at the final Down-sampling and initial Up-sampling stages of U-net, where textural information is established \cite{hung2019image}. This multi-stage approach integrates visual, textual, and grounding information within the T2I model.

\paragraph{\textbf{Context Enhancement:}}
To mitigate the adverse effects of perception features lacking global context on diffusion models, we introduce a Transformer-based feature fusion module, \textit{Enh-Fusion}. Specifically, we concatenate the multi-scale feature from the FPN, with resolutions of $[256 \times 64 \times 64, 256 \times 32 \times 32, 256 \times 16 \times 16, 256 \times 8 \times 8, 256 \times 4 \times 4]$, to obtain a high-dimensional feature sequence of length $256 \times 5231$. To maintain computational efficiency, this sequence is projected into a $256 \times 786$ through a linear transformation. A learnable positional embedding is then added to the transformed features, which are subsequently input into the \textit{Enh-Fusion}. This design enables the perception features to incorporate global contextual information, thereby further enhancing the performance of the diffusion model.

\paragraph{\textbf{Perception Network:}}
The perceptual network comprises a Classification Subnet and a Box Regression Subnet, each implemented using four 3$\times$3 convolutional layers. The Classification Subnet outputs probabilities indicating object presence at each spatial location for three anchors. In parallel, the Box Regression Subnet provides linear outputs of 4$\times$3, representing a bounding box for each spatial location. During training, the perceptual module contributes to the overall loss, thus increasing the T2I model’s sensitivity to object positioning.
\begin{figure*}[!ht]
    \centering
    \includegraphics[width=1\linewidth]{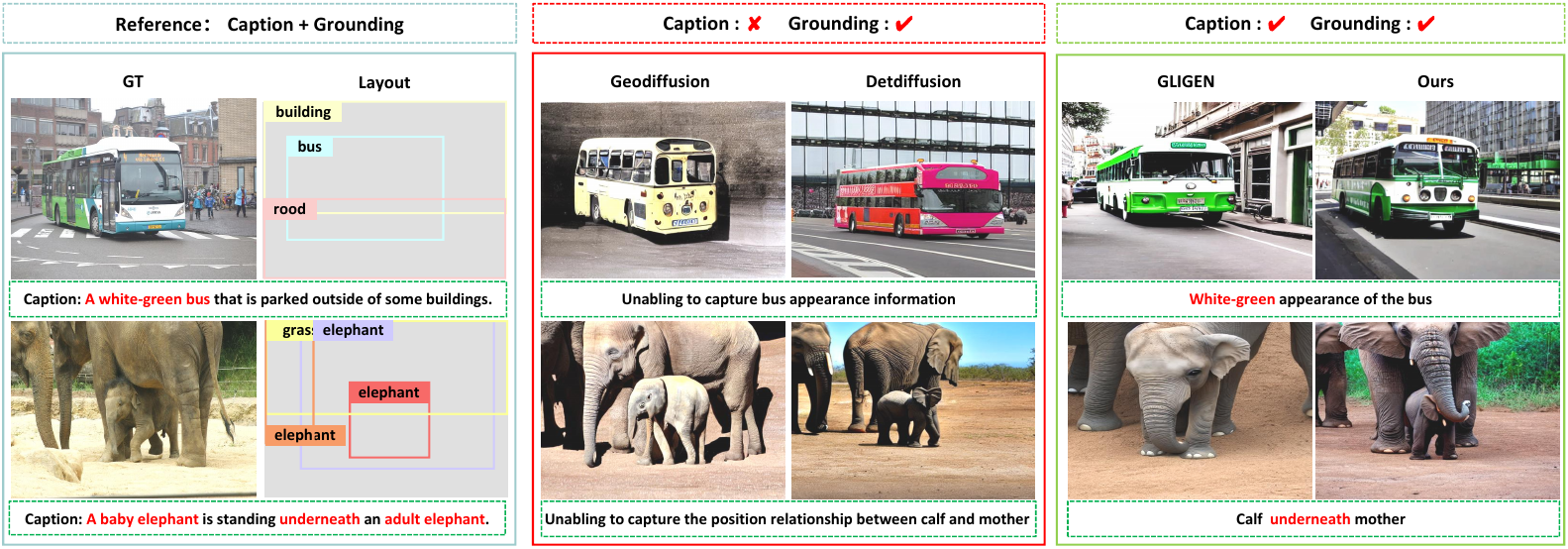}
    \caption{Show the importance of text-captions in generated images, significantly influencing the style, appearance, and layout of the object entities.}
    \label{fig:fig5}
\end{figure*}
\subsection{Main Results}
For Text-to-Image (T2I) tasked with generating objects at precise locations \cite{cheng2023layoutdiffuse,gupta2021layouttransformer}, objects must not only be rendered in their intended spatial positions but also maintain stylistic consistency with the original scene. To demonstrate the efficacy of our approach, we compare our model’s performance with other state-of-the-art T2I models using Fréchet Inception Distance (FID) and mean Average Precision (mAP) metrics. This assessment was conducted on the validation set of the MSCOCO dataset.


\begin{table}[ht]
\centering
\resizebox{0.9\linewidth}{!}{
\begin{tabular}{c|c|ccc|c|c}
\toprule
\multirow{2}{*}{Method} & \multirow{2}{*}{FID $\downarrow$} & \multirow{2}{*}{mAP $\uparrow$} & \multirow{2}{*}{$\mathrm{AP}_{50}$ $\uparrow$} & \multirow{2}{*}{$\mathrm{AP}_{75}$ $\uparrow$} & \multicolumn{2}{c}{Infer (s) $\downarrow$} \\
\cline{6-7}
 & & & & & V100 & 3090~TI \\
\midrule
RB-LostGAN & 42.55 & 9.1 & 15.9 & 9.8 & - & -\\
LAMA & 31.12 & 13.4 & 19.7 & 14.9 & - & - \\
\hline
L.Diffuse$^{\dagger}$ & 22.20 & 11.4 & 23.1 & 10.1 & 2.3 & 4.4 \\
L.Diffusion$^{\dagger}$ & 22.65 & 14.9 & 27.5 & 14.9 & 2.7 & 4.6 \\
ReCo$^{\dagger}$ & 29.69 & 18.8 & 33.5 & 19.7 & 2.2 & 4.1 \\
GLIGEN & 21.04 & 22.4 & 36.5 & 24.1 & 2.8 & 4.6 \\
ControlNet$^{\dagger}$ & 20.37 & 24.8 & 36.6 & 27.7 & 2.6 & 3.8 \\
GeoDiffusion & 20.16 & 29.1 & 38.9 & 33.6 & \cellcolor{green!10}\textbf{1.8} & \cellcolor{green!10}\textbf{3.1} \\
DetDiffusion & 19.66 & 31.2 & 40.2 & 35.6 & 1.9 & 3.3 \\
\midrule
\cellcolor{green!10}\textbf{SpatialLock} &
\cellcolor{green!10}\textbf{18.32} &
\cellcolor{green!10}\textbf{32.1} &
\cellcolor{green!10}\textbf{41.1} &
\cellcolor{green!10}\textbf{36.7} &
2.1 & 3.5 \\
\bottomrule
\end{tabular}
}
\caption{Our approach achieves a clear performance advantage: the FID is \textbf{1.34 points} better and the mAP is \textbf{0.9 points} higher than the previous best model, while maintaining competitive inference speed. This demonstrates the strong generative capability of our model.}
\label{Tab1}
\end{table}


\vspace{-2em}
\subsubsection{Quantitative Evaluation} 
\paragraph{\textbf{mAP Score.}} A higher mean Average Precision (mAP) score indicates greater accuracy in the positioning of generated objects relative to predefined locations. As demonstrated in Table 1, our method achieves state-of-the-art performance, attaining the maximum mAP score. Our results surpass those of DetDiffusion \cite{wang2024DetDiffusion} and GeoDiffusion \cite{chen2023geodiffusion} by $\textbf{0.9}$ and $\textbf{2.0}$ points, respectively. This advantage is attributable to our approach of directly incorporating grounding information into the T2I generative network via the \textit{PoI} module, thereby endowing the model with a stronger capability to generate objects in their anticipated positions. Furthermore, our comparison with GLIGEN \cite{li2023GLIGEN} demonstrates that the inclusion of external position information through \textit{PoG} strengthens the T2I model’s ability to generate images with specific spatial arrangements. These findings suggest a \textbf{\textit{mutually reinforcing interaction between position information and image generation.}}

\paragraph{\textbf{Fidelity.}} A lower FID score signifies that the generated images are more realistic. Similarly to mAP, Our model demonstrates superior image synthesis capabilities compared to existing methods, achieving the lowest FID score. Since utilizing only two cross-attention modules, our model’s trained parameters are less than \textbf{50.7\%} of those in GLIGEN \cite{li2023GLIGEN}, yet it achieves a \textbf{2.72} point improvement in FID. This enhancement is attributable to the injection of richer knowledge through the \textit{PoG }during the training, which provides the T2I generative network with greater oversight. Moreover, with a richer input of information (captions and grounding data), our model outperforms GeoDiffusion and DetDiffusion in FID score by \textbf{1.84} and \textbf{1.34} points, respectively. These results illustrate that abundant position information enhances image generation quality.

\paragraph{\textbf{Inference Efficiency.}} Inference efficiency is a critical factor in assessing the practicality of the method. As shown in Table 1, \textit{SpatialLock} achieves significant improvements in both FID and mAP while maintaining competitive inference speed. Although it is slightly slower than the fastest method, GeoDiffusion, $0.3$s and $0.4$s on V$100$ and $3090$TI GPUs, it outperforms most other counterparts. This advantage stems from \textit{SpatialLock} integrating the Ground-Attention into selected layers of the U-Net, which effectively enhances performance without substantially increasing the network's parameters.

\begin{figure}[h]
    \centering
    \includegraphics[width=1\linewidth]{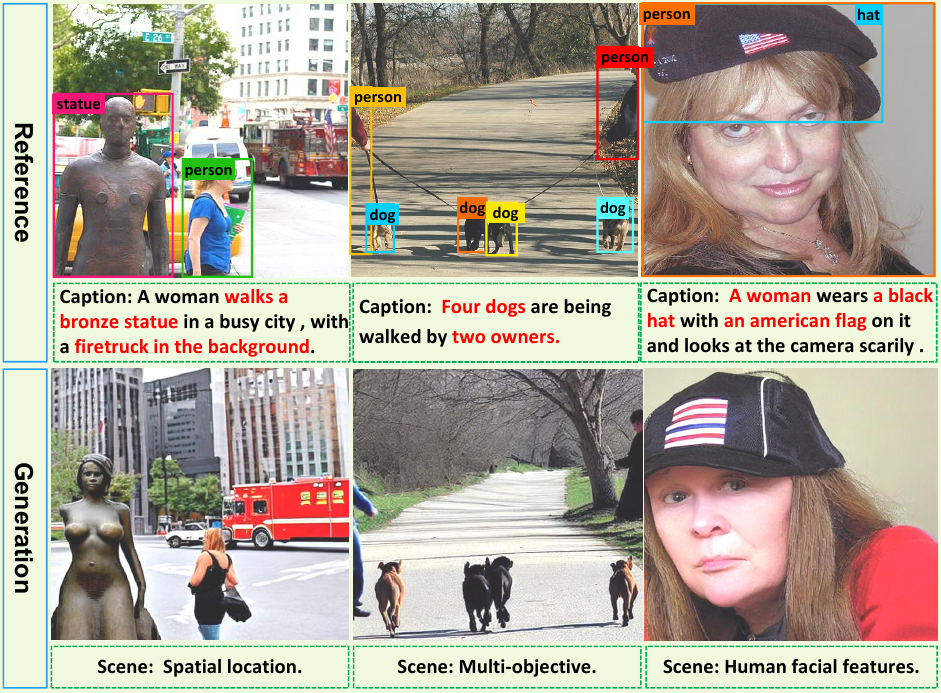}
    \caption{The capability of our model to generate various scenes on the Flickr dataset, including spatial position, Multi-objective generation, and detail of human facial features.}
    \label{fig:fig6}
\end{figure}

\subsubsection{Qualitative Evaluation}
\paragraph{\textbf{Position Accuracy.}} As illustrated in \hyperref[fig:fig4]{Figure 4}, our model demonstrates superior accuracy in object positioning compared to alternative methods. ControlNet \cite{zhang2023adding} exhibits limited position accuracy, primarily due to insufficient learning of spatial information. While DetDiffusion incorporates perception information for supervision, its exclusion of image descriptions hinders the model’s capacity to learn global layouts, thus degrading generation results. And GeoDiffusion faces a similar issue. Our approach overcomes these challenges by implementing \textit{PoI} and \textit{PoG}, leading to improved object position accuracy. We accurately generate the positions of two deer, and demonstrate high precision in scenes with complex positional relationships, such as a ‘white dog’ ‘standing’ on a bench. Especially for small object generation that requires a deep understanding of object layout, such as 'holding a glove', our network still has an excellent performance, proving the two components are complementary.

\paragraph{\textbf{Fidelity.}} Under identical hyperparameter configurations, our model generates more vivid and diverse visual outputs better. As shown in \hyperref[fig:fig4]{Figure 4,} the outputs from ControlNet \cite{yang2023reco} exhibit incompleteness, for instance, a deer with five legs. This limitation arises from ControlNet's inadequate capacity to reconcile the relationship between generated content and position encoding. DetDiffusion, while focusing on the accuracy of generated objects by directly using grounding information as input, overlooks the importance of learning image descriptions, leading to a degradation of detail. GeoDiffusion similarly faces limitations due to its reliance solely on grounding information as input, resulting in diminished image fidelity. Our model effectively mitigates these shortcomings by integrating novel modules that facilitate additional information injection. The generated results exhibit a high degree of congruence with the original scene, and the details of the objects are significantly enhanced. For example, the generated instances of the deer are highly realistic, the details of the ballplayer and glove are even more impressive, the color of the puppy is accurate, and the integrity of the bench is excellent.

\begin{figure}[!th]
    \centering
    \includegraphics[width=1\linewidth]{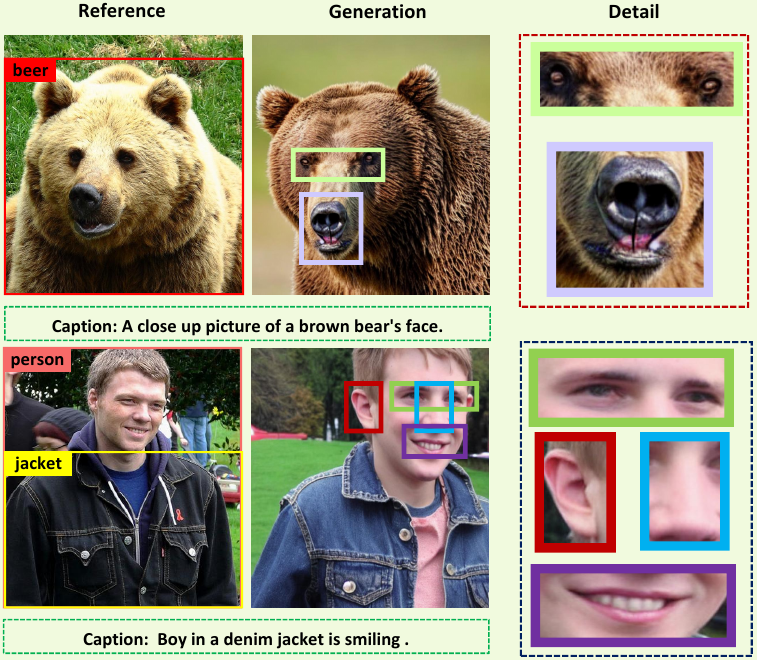}
    \caption{Showcase the capability of our models for generating detail. The facial features of both animals and humans are very vivid.}
    \label{fig:fig7}
\end{figure}

\paragraph{\textbf{Text-Caption Guidance.}} Some researchers discard text-captions, utilizing grounding information directly as input to the model \cite{wang2024DetDiffusion,chen2023geodiffusion}. However, this approach tends to produce unfavorable results. As illustrated in \hyperref[fig:fig5]{Figure 5}, the absence of caption guidance results in a lack of control over the generated results. For instance, GeoDiffusion \cite{chen2023geodiffusion} exhibits randomized backgrounds, DetDiffusion \cite{wang2024DetDiffusion} fails to accurately capture the bus’s color and appearance, and the spatial arrangement of the elephant ‘underneath the adult elephant’ is incorrect. In contrast, GLIGEN \cite{li2023GLIGEN}, which incorporates text-captions, achieves superior performance. Although the generative abilities are somewhat limited, the results still manage to capture key attributes such as 'underneath' and 'white-green'. Compared with GLIGEN, our model enhances comprehensive generative capacity, significantly improving the FID and mAP. These results highlight that a richer information input stimulates the model’s potential to fully integrate these features, subsequently enhancing its capacity for learning object positions.

\begin{table}[!h]
\centering
 \resizebox{1\linewidth}{!}
 {
\begin{tabular}{ccc|c|ccc}
\toprule
PoI & PoG & PoG$_{\mathrm{Enh\_Fusion}}$  & FID $\downarrow$ & mAP $\uparrow$ & AP$_{50}$ $\uparrow$ & AP$_{75}$ $\uparrow$ \\
\midrule
\ding{51} & \ding{55} & \ding{55}  & 22.8 & 21.1 & 35.5 & 20.6 \\
\ding{55} &  \ding{51} & &  28.6 & 11.8 & 21.3 &  14.1 \\
 \ding{55} &  & \ding{51}   & 27.5 &  12.9 &  22.5 &  15.2 \\
\ding{51} &  \ding{51} & &  18.4 &  31.9 &  40.8 &  36.1 \\
\cellcolor{green!10} \ding{51}  & \cellcolor{green!10}  & \cellcolor{green!10} \ding{51}  & \cellcolor{green!10} \textbf{18.32} & \cellcolor{green!10} \textbf{32.1} & \cellcolor{green!10} \textbf{41.1} & \cellcolor{green!10} \textbf{36.7} \\
\bottomrule
\end{tabular}
}
\caption{Compare the impact of different components for our model. Under the joint contribution of the \textit{PoI} and \textit{PoG} equipped with the \textit{Enh-Fusion}, the model achieves its best performance}
\label{Tab2}
\end{table}

\paragraph{\textbf{Flickr Results.}} While the Flickr dataset presents scenes of greater intricacy than those in MSCOCO, our model consistently delivers compelling performance. \hyperref[fig:fig6]{Figure 6} presents diverse generated samples, confirming our model’s ability to synthesize high-fidelity images with accurate spatial layouts.

\paragraph{\textbf{Generation Detail.}} We selected the facial features of both animals and humans to showcase our model’s capacity for generating intricate details, as facial feature synthesis is inherently challenging. As depicted in \hyperref[fig:fig7]{Figure 7}, the generated eyes and noses of brown bears exhibit striking realism, free from mosaic-like. Furthermore, the generated human facial features are enough to confuse the real with the fake, which highlights the exceptional generative power of our network.

\subsection{Ablation Studies}
\paragraph{\textbf{Component Ablation.}} By sequentially integrating the components into the network, we gradually verify the impact of \textit{PoI} and the \textit{PoG} on our model's performance. As Table 2 indicates, we first integrate the \textit{PoI}, which introduces position information via Grounding-Tokens, allowing the model to directly acquire positional features. This results in a significant enhancement in object position accuracy. Subsequently, the \textit{PoG} provides external supervision, enabling the T2I model to learn more comprehensive position information and further improve object positional accuracy. The supplementary position information prompts the Grounding-Attention modules to better explore textural characteristics, thereby augmenting the fidelity of the generated images. Meanwhile, the \textit{Enh-Fusion} module further improves the overall performance of the network, which, due to its superior capability in contextual feature integration, compensates for the lack of global context information in the perception module.



\begin{table}[h]
\centering
\resizebox{1\linewidth}{!}{
\begin{tabular}{cccccc|c|c|c}
\toprule
$D_1$ & $D_2$ & $D_3$ & $U_1$ & $U_2$ & $U_3$ & FID $\downarrow$ & mAP $\uparrow$ & Infer (s) \\
\midrule
\ding{51} & \ding{51} & \ding{51} & \ding{51} & \ding{51} & \ding{51} & 22.9 & 30.1 & 3.2 \\
\ding{51} & \ding{51} & \ding{51} & \ding{55} & \ding{55} & \ding{55} & 32.4 & 21.8 & 2.3 \\
\ding{55} & \ding{55} & \ding{55} & \ding{51} & \ding{51} & \ding{51} & 29.7 & 22.9 & 2.4 \\
\ding{55} & \ding{51} & \ding{51} & \ding{51} & \ding{51} & \ding{55} & 20.4 & 31.3 & 2.7 \\
\rowcolor{green!10}
\ding{55} & \ding{55} & \ding{51} & \ding{51} & \ding{55} & \ding{55} & \textbf{18.3} & \textbf{32.1} & \textbf{2.1} \\
\bottomrule
\end{tabular}
}
\caption{Injecting grounding information into the semantic-rich layers of the U-Net allows for better coordination between perception and grounding information.}
\label{Tab3}
\end{table}


\vspace{-3.em}
\paragraph{\textbf{Grounding in U-Net.}} We inserted the Ground-Attention at different layers of the U-Net to explore its impact on performance. As shown in Table 3, ($D\_x$ and $U\_x$ represent the Down-sampling and Up-sampling in the U-Net, respectively), introducing position information into all layers of U-Net resulted in performance degradation. We attribute this to the excessive accumulation of position information that overwhelms the generative process and acts as noise, thereby weakening the effectiveness of perceptuion guidance. When grounding information is injected only during either the Down-sampling or Up-sampling stages, the asymmetric architecture tends to underperform. This is primarily because both the encoder and decoder require access to global contextual information during the generation process. In contrast, when the Ground-Attention is applied only at the last Down-sampling and the first Up-sampling layers, the position and perception information can work more synergistically, leading to the best performance. These results indicate that selectively injecting grounding information into the network is crucial for achieving optimal performance. 


\subsection{Additional Exploration}
\paragraph{\textbf{Robust Generalization.}} To validate the generalization capabilities of our model across diverse T2I datasets, we conducted a cross-dataset evaluation. For example, we trained our model on MSCOCO and tested it on Flickr. As shown in Table 4, all experiments achieved reliable FID and mAP scores. These results demonstrate our network’s capacity to generate novel scenes effectively. Furthermore, with richer input information, the robustness of our T2I generative capacity is significantly enhanced.

\begin{table}[h!]
\centering
\begin{tabular}{c|c|c|cc}
\toprule
Train Dataset & Val Dataset & FID  & mAP  \\
\midrule
MSCOCO & Flcker & 24.3 & 27.2   \\
MSCOCO & NuScenes & 26.7 & 24.6  \\
Flicker & MSCOCO & 22.2 & 29.1  \\
Flicker & NuScenes & 25.1 & 26.1  \\
\bottomrule
\end{tabular}
\caption{Demonstrate our model's ability for robust generalization in different value sets.}
\label{Tab4}
\end{table}

\vspace{-2.5em}
\paragraph{\textbf{MSCOCO Expansion.}} A total of $20,000$ synthetic images with precision position, styled according to the MSCOCO dataset, were generated by \textit{SpatialLock} and subsequently added to the original training set. The mainstream object detection networks were then retrained using this augmented dataset. As shown in Table 5, after incorporating the new synthetic data, the performance of all models improved to varying degrees. This shows that our method effectively addresses the challenges associated with dataset acquisition and annotation.

\begin{table}[h]
\centering
\begin{tabular}{c|c|c|c}
  \toprule
  \multirow{2}{*}{Detector}  & \multicolumn{2}{c|}{mAP} & \multirow{2}{*}{UP} \\
  \cline{2-3}
                  & MSCOCO  & MSCOCO+Gen   \\
  \hline
    SSD300 \cite{liu2016ssd}       & $28.0$   & $30.1$  &   \cellcolor{green!10} \textbf{2.1} \\
    RetinaNet \cite{lin2017focal}   & $32.5$   & $35.7$  &   \cellcolor{green!10} \textbf{3.2} \\
    Fast-Rcnn \cite{ren2016faster}   & $35.9$   & $37.5$  &   \cellcolor{green!10} \textbf{1.6} \\
    Fcos   \cite{tian2019fcos}     & $37.1$   & $40.3$  &   \cellcolor{green!10} \textbf{3.2} \\
    Yolov4  \cite{bochkovskiy2020yolov4}    & $42.4$   & $44.5$  &   \cellcolor{green!10} \textbf{2.1} \\
\bottomrule
\end{tabular}
\caption{The performance of the classical object detection algorithm is improved after retraining on MSCOCO+Gen.}
\label{Tab5}
\end{table}

\vspace{-2em}
\section{Declaration}
This research was conducted in accordance with the highest ethical standards for AI research. We utilized two publicly available datasets, MSCOCO and Flickr, which do not contain any personally identifiable information, ensuring that no privacy concerns arise. Furthermore, all methodological strategies presented in this paper are novel and free from plagiarism, fabrication, or other forms of academic misconduct. Finally, we affirm that this work contains no elements of racial, ethnic, or regional discrimination.

\section{Conclusion}
For generating images with precise positional control, we introduce a novel framework named \textit{SpatialLock}, which first utilizes grounding and perceptual information for the joint supervision of position learning. The framework incorporates two modules: Position-Engaged Injection (\textit{PoI}) and Position-Guided Learning (\textit{PoG}). \textit{PoI} employs a specifically designed attention layer to encode grounding signals, while \textit{PoG} leverages perception information to provide additional supervision. This unique design enriches the input information, enabling \textit{SpatialLock} to achieve state-of-the-art performance in generating images with accurate spatial layouts.

\bibliographystyle{ACM-Reference-Format}
\bibliography{sample-base}

\appendix
\newpage
\newpage
\clearpage
This supplementary material outlines the foundational generative models, clarifies the differences of our method with others, describes the influence of hyperparameters, and showcases further generated results.

\section{Stable Diffusion Model}
Stable Diffusion (SD) model \cite{ramesh2022hierarchical,saharia2022photorealistic,rombach2022high},  recognized as an efficient implementation of Latent Diffusion Models (LDMs), has become a prevalent paradigm for Text-to-Image (T2I) tasks. SD leverages a pre-trained Variational Autoencoder (VAE)  \cite{wang2016auto} to encode an input image $\boldsymbol{x}$ into a latent space feature representation $\boldsymbol{f}$. This latent representation significantly reduces the dimensionality compared to the original image. 
Currently, a Contrastive Language Image Pretraining Model (CLIP) \cite{radford2021learning} encodes the input text-captions into a feature vector $\boldsymbol {c}$. During the forward diffusion process, random noise $\boldsymbol{n}$ is gradually added to the latent feature $\boldsymbol{f}$ over a sequence of timesteps $\boldsymbol{t}$. Subsequently, the caption embedding $\boldsymbol{c}$ and the latent feature $\boldsymbol{f}$ are integrated into a U-Net network \cite{ronneberger2015u}, which integrates the CNN and Transformer layers. The CNN layers refine the latent space features, while the Transformer layers primarily facilitate the fusion of caption and visual information. This integrated architecture is the key to the superior generative capabilities of the model. The formulation of the objective function can be expressed as follows:
\begin{equation}
    \begin{aligned}
        L_{SD}=E_{\xi{(x)},\epsilon \sim N(0,1),t}\Vert \epsilon-\epsilon_\theta (z_t,t,\tau_{\theta}(y) \Vert^2
    \end{aligned}
\end{equation}
 Where ${t}$ is the sampling of timesteps, ${z}_t$ is the noisy variant of input ${z}$ in step ${t}$, ${\epsilon_\theta (z_t,t,\tau_{\theta}(y))}$ is the ${(t, \tau_{\theta}(y))}$-conditioned denoising of the autoencoder. SD models primarily predict the similarity between the added and output noise, enabling the generative models to be equipped with the ability to recognize the additional noise.

\begin{table}[!h]
\centering
\begin{tabular}{cc|cccc}
\toprule
$\boldsymbol{\alpha}$ & $\boldsymbol{\beta}$ &
\textbf{FID} $\downarrow$ & \textbf{mAP} $\uparrow$ &
$\mathbf{AP}_{50}$ $\uparrow$ & $\mathbf{AP}_{75}$ $\uparrow$ \\
\midrule
1.0 & 0.0 & 27.5 & 12.9 & 22.5 & 15.2 \\
0.9 & 0.1 & 26.9 & 14.8 & 22.1 & 15.7 \\
0.5 & 0.5 & 23.9 & 18.2 & 25.3 & 22.9 \\
0.1 & 0.9 & 24.5 & 28.9 & 25.5 & 23.7 \\
0.0 & 1.0 & 22.8 & 21.1 & 35.5 & 24.6 \\
\hline
\rowcolor{green!10}
\textbf{Trainable} & \textbf{Trainable} &
\textbf{18.32} & \textbf{32.1} & \textbf{41.1} &  \textbf{36.7} \\
\bottomrule
\end{tabular}
\caption{The impact of $\alpha$ and $\beta$ on the performance of our model.}
\label{Tab6}
\end{table}


\vspace{-2em}
\section{Influence by Hyperparameter.} As shown in Table 6, we investigated the impact of perception and generation loss coefficients $\alpha$ and $\beta$ on our network’s performance. We designed several different fixed values, but none of them led to satisfactory performance. This is because, in diffusion models, perceptual and generative information represent a trade-off that needs to be balanced, and the network must learn the relationship between them automatically.

\section{Grounding-Attention vs Gate-Attention}
The Gate-Attention mechanism in GLIGEN \cite{li2023GLIGEN}, in processing visual and grounding information, typically encodes both modalities into a unified representation, subsequently fusing them via a self-attention layer. Although this approach offers simplicity and directness, it suffers from inherent limitations. Specifically, the rich semantic detail in visual information potentially subsumes the relatively sparse location feature. This results in location features lacking sufficient purity, hindering the model’s capacity to generate images with precise location. 

\begin{figure}[!h]
    \centering
    \includegraphics[width=0.75\linewidth]{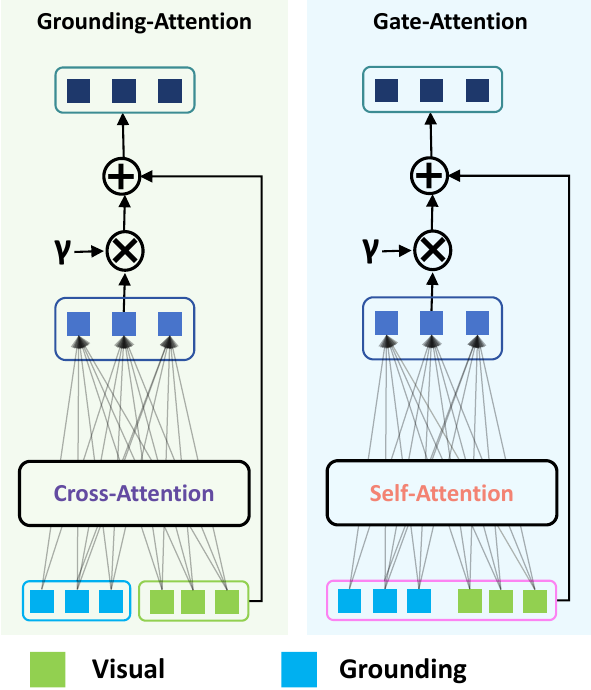}
    \caption{Highlights the similarities and differences between Grounding-Attention and Gate-Attention. Unlike GLIGEN, which treats visual and grounding information jointly, our approach models them as independent entities, capturing their relationship via cross-attention.}
    \label{fig:fig8}
\end{figure}

As shown in \hyperref[fig:fig8]{Figure 8}, to mitigate the limitations, we introduce a novel methodology that decouples the processing of visual and location features. In contrast to holistic fusion, our approach treats visual and grounding features as independent entities. The core principle underlying this strategy is to minimize the influence of visual information on location features, thereby ensuring that the model learns a location-centric representation. 

Our proposed approach, by independently processing location features, guides it to focus predominantly on critical spatial attributes. Mitigating the interference from visual information, thus yielding a more refined representation of spatial context. This enhanced purity of location features contributes to increased precision in location and mitigates potential information conflation, enabling the model to leverage location information more effectively across various visual tasks. These pure location features benefit the perception module in learning location information and contribute to improved supervisory efficacy of the module.

\begin{figure*}[!h]
    \centering
    \includegraphics[width=1\linewidth]{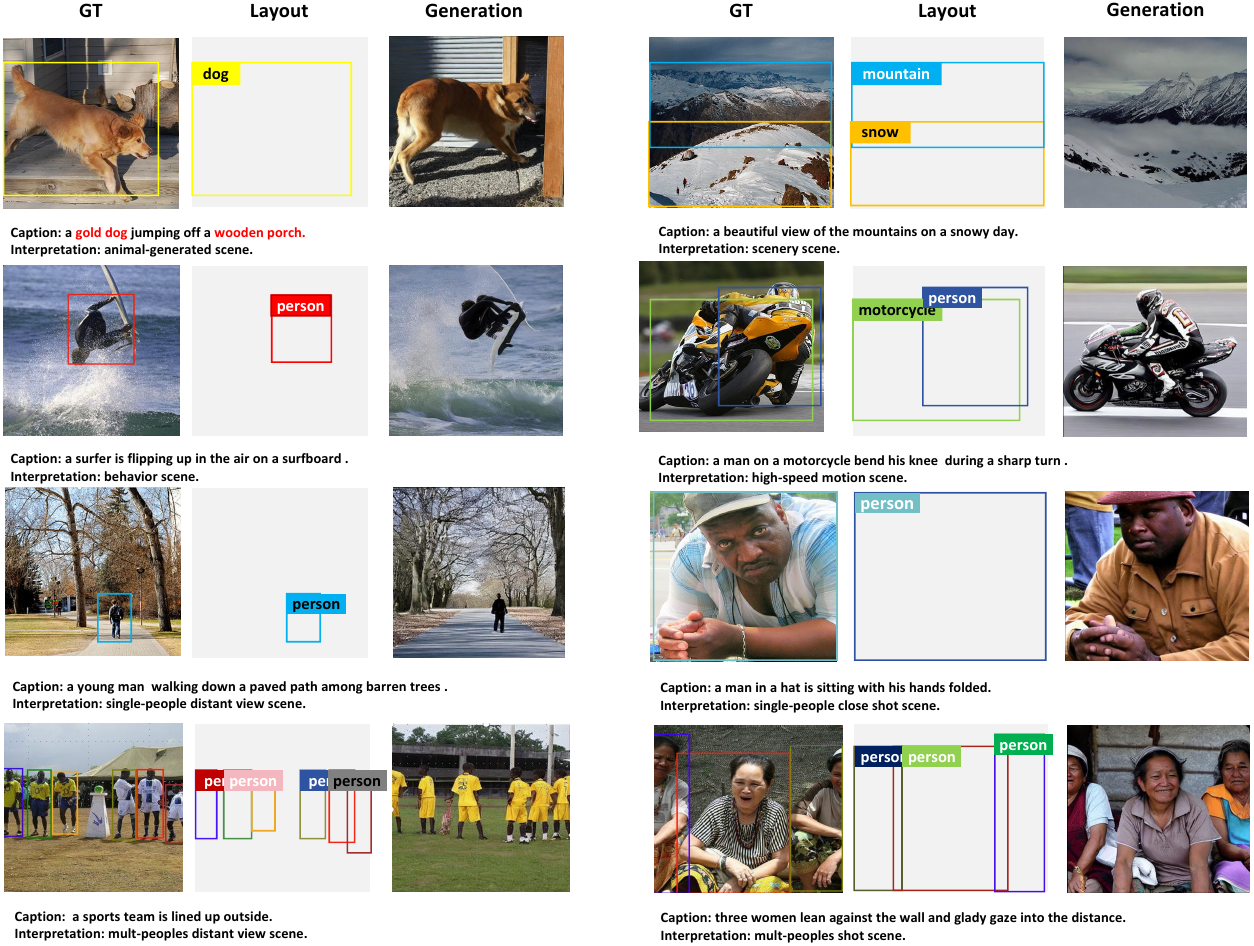}
    \caption{Shows model’s generative capabilities across diverse scenes on Flickr, showcasing its effectiveness in a variety of complex scenarios.}
    \label{fig:fig9}
\end{figure*}

\section{More Results.} As shown in \hyperref[fig:fig9]{Figure 9}, the Flickr dataset offers more complex and realistic scene settings compared to MSCOCO. When we tested our model on Flickr, \textit{SpatialLock} continued to perform exceptionally well.

\end{document}